\documentclass[review]{elsarticle}

\usepackage{hyperref}
\usepackage{amsmath}
\usepackage{booktabs}
\usepackage{url}
\usepackage{microtype}

\DeclareMathOperator{\sign}{sign}

\journal{Neurocomputing}

\begin{document}

\begin{frontmatter}

\title{An Approximate Backpropagation Learning Rule for Memristor Based Neural Networks Using Synaptic Plasticity}

\author[mipt]{D.~Negrov\corref{cor1}}
\ead{dmitriynegrov@gmail.com}
\cortext[cor1]{Corresponding author}
\author[mipt,srisa]{I.~Karandashev}
\author[mipt,srisa]{V.~Shakirov}
\author[mipt,mephi]{Yu.~Matveyev}
\author[mipt,srisa]{W.~Dunin-Barkowski}
\author[mipt,mephi]{A.~Zenkevich}

\address[mipt]{Laboratory of Functional Materials and Devices for Nanoelectronics, Moscow Institute of Physics and Technology, Dolgoprudny, Russian Federation}
\address[srisa]{Scientific Research Institute for System Analysis, Russian Academy of Sciences, Moscow, Russian Federation}
\address[mephi]{National Research Nuclear University “MEPhI”, Moscow, Russian Federation}

\begin{abstract}
We describe an approximation to backpropagation algorithm for training deep neural networks, which is designed to work with synapses implemented with memristors. The key idea is to represent the values of both the input signal and the backpropagated delta value with a series of pulses that trigger multiple positive or negative updates of the synaptic weight, and to use the min operation instead of the product of the two signals. In computational simulations, we show that the proposed approximation to backpropagation is well converged and may be suitable for memristor implementations of multilayer neural networks.
\end{abstract}


\end{frontmatter}

\section{Introduction}

Recent advances in machine learning have provided the solutions to many problems, which seemed insurmountable in the past. New approaches based on deep and recurrent artificial neural networks (ANN) are very efficient in dealing with pattern recognition, visual objects detection, speech recognition, signal restoration and prediction, etc. \citep{LeCun2015}. Deep feed forward networks excel at these tasks because their design can be general enough to solve many natural classification tasks, but their main feature is a funnel-like error landscape, which opens a possibility to apply computationally efficient learning methods, such as stochastic gradient descent.

Nowadays, the computations with ANNs are performed mostly using conventional computation architectures by simulation on general purpose processors or graphics cards which are quite efficient at matrix algebra being the core of feed forward deep networks. However, due to the limited level of parallelism this approach is quite inefficient in terms of the ratio between the computational speed and the power consumption. Consequently, it hinders the application of these methods in areas, where the available power is limited, such as mobile devices or autonomous robotic platforms. A further step in applying deep learning neural networks to real-life problems would depend on their implementation in hardware which would use accelerators specifically built to perform neural computations and employ the most possible parallelism. The optional solution here is to create a system which mimics the structure of the network under consideration by implementing neurons and their interconnections directly in hardware. In this case, the computational fabric of the device can be made of blocks representing neurons, which are interconnected with synapses with weights implemented as distributed memory, similar to the biological neural tissue.

One of the major problems hindering the design of artificial neural network hardware is the storage of synaptic weights. The distributed approach to the storage makes use of dynamic RAM (DRAM) very inconvenient due to the requirement of constant refresh operations arising from the leakage of charge \citep{ Adhikari2015}.

Among the approaches existing today, the one coming close to take full advantage of huge parallelism of ANNs is an FPGA-based implementation, such as described in  \citep{Zhang2015}. Such implementations rely on the array of specialized processing units, tailored to compute outputs of neural network layer. Being quite efficient, this approach, however, suffers from two drawbacks. Firstly, the distributed RAM on modern FPGA is implemented using static RAM arrays, which have low density, making the available memory limited. Hence, in this case, the implementation of large-scale deep network might require constant exchange with the external DRAM, which is inconvenient. Secondly, despite being quite universal, FPGAs suffer from low connectivity and limited availability of multiply-and-accumulate modules, and therefore they must be shared among different neurons. Moreover, the incorporation of on-line learning in such devices generally requires supervision from the external CPU.

Over the last several years, a variety of memristive devices has been discovered. These devices are the resistors with the conductance controlled by the current or voltage previously applied to them. In this way, they represent the class of non-volatile memory devices with effective continuum of memory states, which can be scaled down to a ten nanometers. These properties of memristors have opened an opportunity to implement a hardware neural network as an array of cores, which consist of a bank of presynaptic and postsynaptic neurons, made with the use of conventional CMOS technology, and interconnected with an array of memristive devices. Such arrays can be arranged as cross-bar arrays, thus providing an efficient storage medium \citep{Likharev2011}, \citep{Govoreanu2011}, \citep{Indiveri2011}, \citep{Indiveri2013}, \citep{Matveyev2015}, useful for the  effective modeling of physiological processes \citep{Wang2016} and the implementation of wide-spread types of neural networks, such as cellular neural networks (CNN) \citep{Hu2015}. The comprehensive survey of memristors and memristor-based techniques with the necessary references to previous works are given in the seminal paper of the field's pioneer L. Chua \citep{Chua2015}. 

For the hardware implementation of hybrid CMOS/memristor neural networks, one of the biggest challenges is training \citep{Merkel2014SPIE}. Even simplest training algorithms, such as error backpropagation \citep{Werbos1975}, are difficult to implement at the circuit level. It is much easier to implement these algorithms off the chip and then transfer the training results \citep{Merkel2015}. Another approach is to train only the output layer of the network, which can be accomplished using a simple least-mean-squares learning algorithm \citep{Alibart2013}, \citep{Merkel2014SPIE}, \citep{Widrow2013}, \citep{Singh2014}.

In \citep{Adhikari2015}, an alternative Random Weight Change learning algorithm is used. Although its convergence is theoretically proven, it appears to be very slow. For each synaptic weight it uses four memristors in a bridge configuration. Besides, there is an independent random number generator in each model neuron. The scheme in \citep{Hasan2014} uses standard backpropagation scheme and two memristors for each synapse. Unfortunately, the learning is implemented there using an external training unit, and this part of the network learning essentially uses MATLAB, because it apparently does not have a simple implementation in circuitry, while the other part of the modeling is performed via SPICE simulator. A significant advantage of the work \citep{Soudry2014} is the clearly formulated problem of \textit{non-local calculation of the signal and error product at the opposite contacts of a memristor}. However, the particular solution proposed by the authors critically depends on the linearity of the memristor resistance as a function of both the amplitude and the duration of the passed current. This condition does not yet hold for the majority of the current memristor implementations.

Meanwhile, there has been a large number of works emulating a so-called synaptic plasticity in memristor devices (e.g., \citep{Jo2012}, \citep{Matveyev2015}). They are aimed at demonstrating effects similar to the plasticity of biological synapses, such as short and long term potentiation/depression and spike-timing-dependent synaptic plasticity. Despite the fact that these models are biologicaly inspired, there is no clear idea of how to use them for the implementation of a learning procedure which would be adequate for solving machine learning problems.

In this paper, we aim to bridge this gap and propose a modification of backpropagation algorithm which allows to circumvent the described difficulties using the learning rule similar to spike-timing-dependent plasticity in synapses. Strictly speaking, we do not give a hardware neural network implementation scheme with a memristor crossbar, but we thoroughly describe the method of pulse representation of signals and the absmin operation used instead of standard product, and also perform the modeling experiments in MATLAB environment.

The remainder of the paper is organized as follows: in Section \ref{ProSt}, we give the problem statement. In Section \ref{Back} the backpropagation algorithm is briefly described. In Section \ref{SigRep} we propose our method of signal representation and the operation used instead of standard product. Section \ref{Exper} describes the test problem of handwritten digits recognition and parameters of modeling experiments. In Section \ref{Results} the results are given. Finally, the discussion of the proposed methods and the obtained results is given in Section \ref{Disc}.

\section{Material and methods}\label{MatMed}

\subsection{Problem statement}\label{ProSt}

In this work, we do not consider any specific scheme of multilayer neural network implementation on a memristor crossbar. However, we assume that a neuron is a device which sums the currents incoming from other neurons. These currents are defined by the conductance of memristors connecting the neurons, i.e. the synapse weight between the two neurons is the conductance of the interconnecting memristor. Thus, in order to set one weight, one memristor is enough, and the memristor crossbar forms the matrix of connections among all neurons.

The problem of neural networks implementation, irrespective of whether it is implemented in software or on the chip, can be divided into two parts: the inference problem and the learning of the network. In this paper, we do not consider the problem of inference, because the forward signal propagation through the multilayer network is reduced to currents passing through memristors crossbar, which is the direct consequence of the Ohm’s law.

The issue considered in this work concerns the learning of the network, where the weight (connection) of synapses between neurons is represented by the memristor conductance. According to the backpropagation algorithm (see Section \ref{SigRep}), the weight update should be proportional to the product of two variables: the forward signal $x_i$ and backward error signal $\delta_j$. The values $x_i$ and $\delta_j$ are calculated in different neurons (Figure~\ref{fig:perceptron}), located in different layers of the neural network. It is desirable for them to interact only via the synaptic connection, and this interaction constitutes the main difficulty for hardware implementation of backpropagation learning rule in multilayer neural networks. In Section \ref{SigRep}, we will show how in this case the signals $x_i$ and $\delta_j$ should be represented in neurons in order to learn the network.

\subsection{Backpropagation}\label{Back} 

In this section, we give a short sketch of the backpropagation technique (the readers familiar with the backpropagation learning rule can skip it and immediately move to the next section). 

\begin{figure}[ht]
	\begin{center}
		\includegraphics[width=8cm]{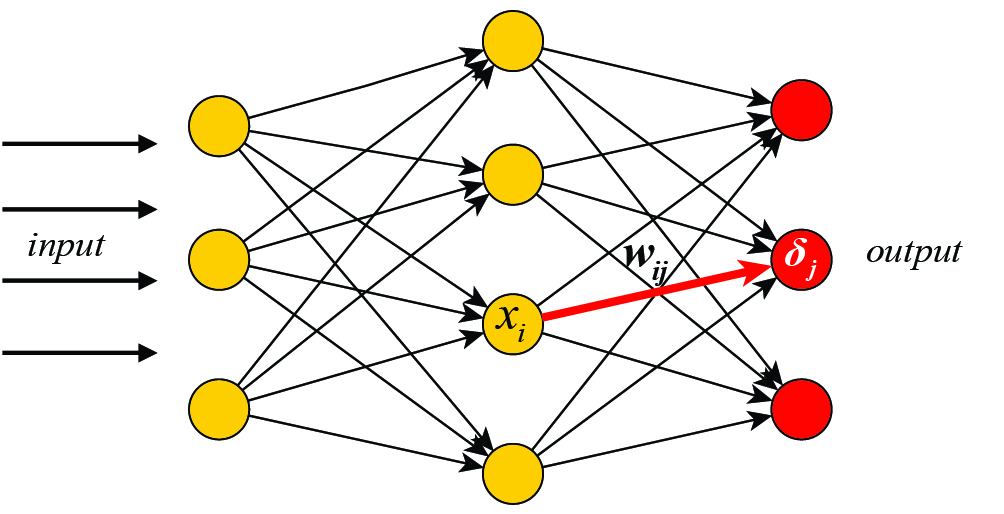}
	\end{center}
	\textbf{\refstepcounter{figure}\label{fig:perceptron} Figure \arabic{figure}.}{ The scheme of a multi-layer perceptron }
\end{figure}

The task of a multilayer perceptron is to get the desired output to the inputs of certain types. With this goal in mind, the learning of perceptron is performed. The pairs of inputs and the desired outputs are loaded to the scheme and the error of the response is determined. The parameters of the scheme (the weights of inter-neuronal connections) are changed with each load, so as to diminish the difference between the desired and the real output (Figure~\ref{fig:perceptron}).

Let us denote the desired output of the neurons as $t_j$, and its actual output as $y_j$. We will further use two measures of network error --- the mean--square--error $\epsilon_{MSE}$ and the cross--entropy $\epsilon_{CE}$:

\begin{equation}
	\epsilon_{MSE} = \sum_j(y_j-t_j)^2\label{eq:crossentropy}
\end{equation}
\begin{equation}
	\epsilon_{CE} = -\sum_jt_j\ln y_j + (1-t_j)\ln(1-y_j)
\end{equation}

Let $w_{ij}$ be the weight of the connection between $j$-th neuron of the last layer and $i$-th neuron of the previous layer (Figure~\ref{fig:perceptron}). The derivative of error with respect to this variable can be written as:

\begin{equation}
	\dfrac{\partial \epsilon}{\partial w_{ij}} = \dfrac{\partial \epsilon}{\partial y_j}\dfrac{\partial y_j}{\partial z_j}\dfrac{\partial z_j}{\partial w_{ij}} = \dot\epsilon(y_j)\dot f(z_j)x_i=\delta_jx_i
	\label{eq:err_decomposition}
\end{equation}
where
\begin{align*}
	z_j&=\sum_iw_{ij}x_i,\\
	y_j&=f(z_j),\\
	\delta_j&=\dot E(y_j)\dot f(z_j)
\end{align*}

Let the output neurons have a sigmoidal activation function $f(z)=\frac{1}{1+\exp(-z)}$, and we use equation (\ref{eq:crossentropy}) for the error calculation. Then, for the equation (\ref{eq:err_decomposition}) we get:

\begin{equation}
	\dfrac{\partial \epsilon}{\partial w_{ij}} = (y_j - t_j)x_i.
	\label{eq:err_last_crossentropy}
\end{equation}

\begin{figure}[ht]
	\begin{center}
		\includegraphics[width=12cm]{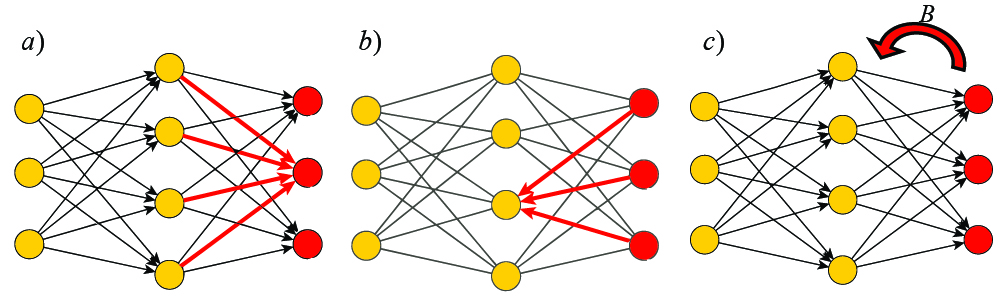}
	\end{center}
	\textbf{\refstepcounter{figure}\label{fig:backprop} Figure \arabic{figure}.}{ Error back-propagation: a) forward propagation of signals; b) error propagation in the backward direction using the same bonds as in a, but in the opposite the direction; c) ....}
\end{figure}

Let us assume that the layers of neurons are numbered from $0$ to $L$, and the weights of connections from the layer $i-1$ to the layer $i$ have the upper index $i$. Then, for all intermediate layers we can obtain expressions similar to (\ref{eq:err_last_crossentropy}).

\begin{equation}
	\dfrac{\partial \epsilon}{\partial w_{ij}^{(k)}} = x_i^{(k-1)}\delta_j^{(k)},
	\label{eq:err_all_layers}
\end{equation}
where $\delta_j^{(L)} \equiv \delta_j = (y_j - t_j)$, $x_i^{0}$ are input signals and
\begin{equation}
	\delta_j^{(k-1)}=\sum_i w_{ji}^{(k)}\delta_i^{(k)}\dot f(z_j^{(k-1)}),
	\label{eq:delta_backprop}
\end{equation}
so that error terms $\delta_j^{(k)}$ are propagated backwards using transposes of weight matrices and the derivative of the activation function. Thus, the back-propagation algorithm is the pair of equations (\ref{eq:err_all_layers}) and (\ref{eq:delta_backprop}) with an additional rule for the weight update. Figure \ref{fig:backprop} shows how signal propagates forward and backward in the network.

Alternatively, the error can backpropagate using another weight matrix $B^{(k)}$, which is different from $W^{(k)}$ and is kept constant during the learning process. Such method was devised in the work \citep{Lillicrap2014}, where elements of matrix $B$ were chosen randomly. Formula (\ref{eq:delta_backprop}) in this case can be rewritten as:
\begin{equation}
	\delta_j^{(k-1)}=\sum_i b_{ji}^{(k)}\delta_i^{(k)}\dot f(z_j^{(k-1)}),
	\label{eq:delta_backprop_alignment}
\end{equation}

\subsection{Signal representation and multiplication using memristors}\label{SigRep} 

Despite the fact that the following procedure is quite general, we are aiming to use it for the implementation of neuromorphic systems employing metal oxide based memristors. This class of memristors has a very non-linear behavior, and their characteristic feature is the presence of a voltage "dead zone", i.e. up to the certain voltage levels, the memristor resistance state is kept unchanged. Moreover, the gradual change of the state in such device can be achieved by applying a voltage pulse train --- the property which has been demonstrated in the works reporting on synaptic plasticity effects \citep{Jo2012}, \citep{Matveyev2015}. This property paves the way for the implementation of a learning procedure in memristor matrices using only local operations as follows. 

In order to use the formula (\ref{eq:err_all_layers}) in optimization algorithms using gradient descent, the weight update should be proportional to the product of two variables:
\begin{equation}
	\Delta w_{ij} \propto\ x_i\delta_j.
	\label{eq:delta_as_product}
\end{equation}

\begin{figure}[ht]
	\begin{center}
		\includegraphics[width=12cm]{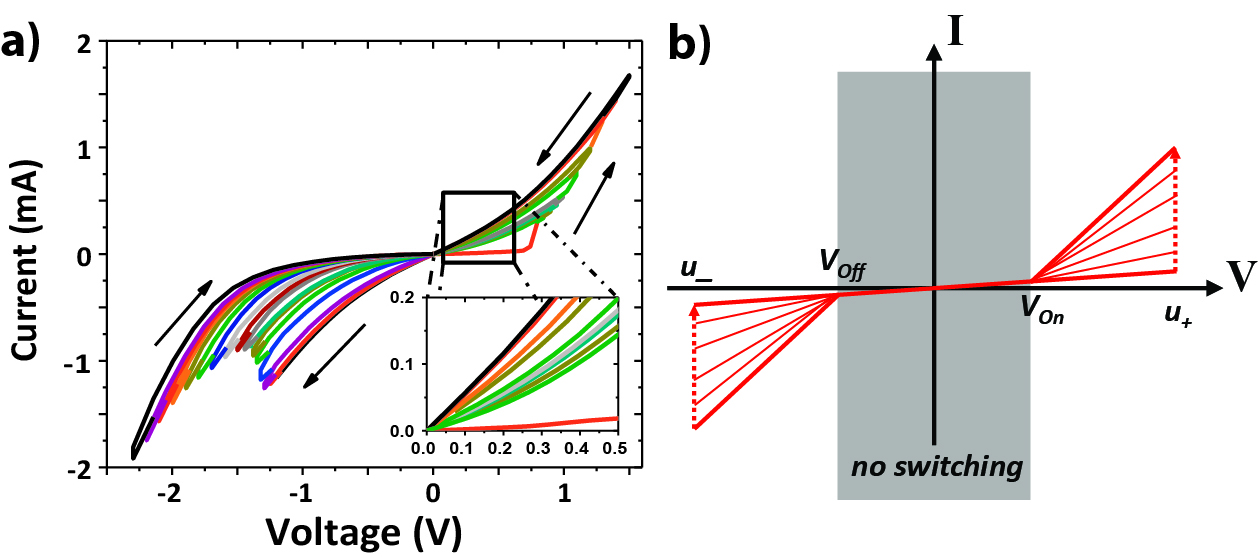}
	\end{center}
	\textbf{\refstepcounter{figure}\label{fig:CVcurve} Figure \arabic{figure}.}{ a) The experimental I-V curves corresponding to the resistive switching cycles of a Pt/Hf$_{x}$Al$_{1-x}$O$_{1-y}$/TiN memristor~\citep{Markeev2013}; b) the schematic drawing of the memristor switching cycles. }
\end{figure}

The values $x_i$ and $\delta_j$ are obtained in different neurons (Figure~\ref{fig:perceptron}), located in different layers of the neural network. It is desirable for them to interact only via the synaptic memristor. To deal with this problem, we use the following approach: taking high non-linearity of metal oxide memristors as an advantage, we choose such a voltage $u_+ > V_{on}$, which results in the change of the memristor state (resistance) (Figure~\ref{fig:CVcurve}), while half of that value does not lead to any changes ($u_+/2 < V_{on}$). By applying voltage pulses with amplitudes $u_+$, one can change the synapse resistivity in small steps. Similarly, we choose voltage pulses of amplitude $u_- <V_{off}$ to change conductivity in the opposite direction.

\begin{figure}[ht]
	\begin{center}
		\includegraphics[width=12cm]{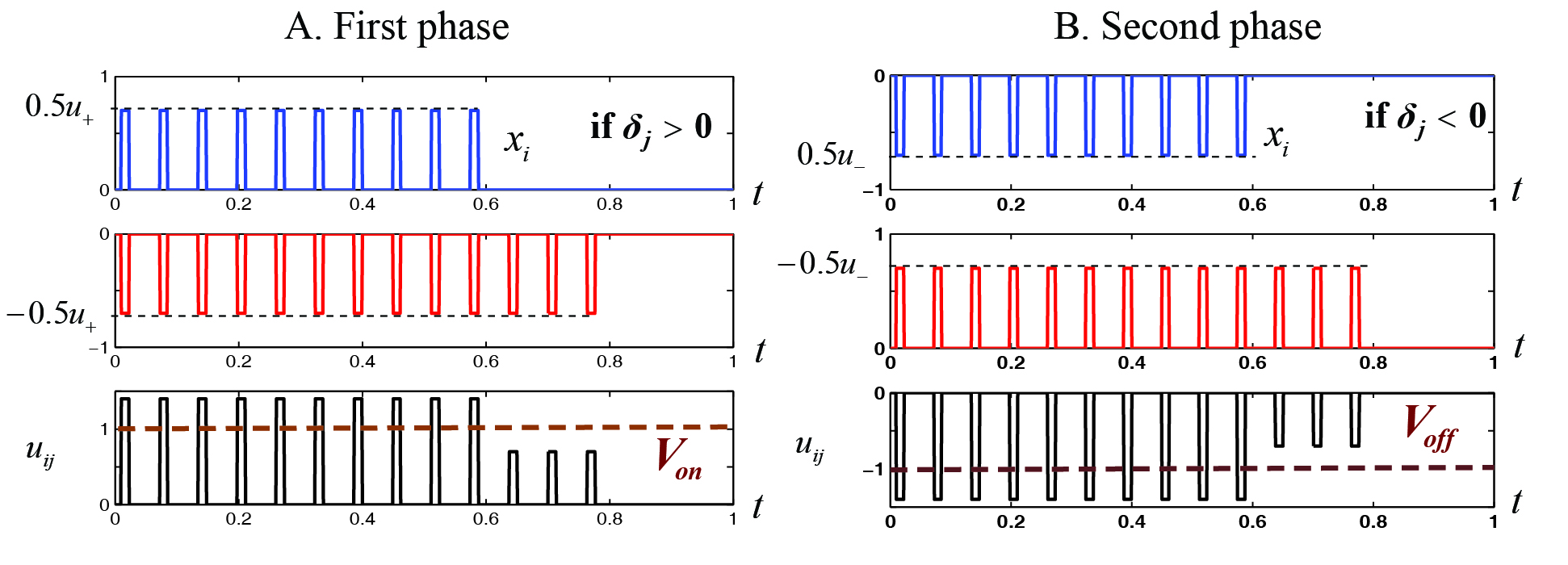}
	\end{center}
	\textbf{\refstepcounter{figure}\label{fig:pulsemodul} Figure \arabic{figure}.}{ Signal waveforms on memristor electrodes: the upper panel illustrates the pulse representation of the signal $x_i$ at one electrode of a memristor, the middle panel illustrates the pulse representation of the signal $\delta_j$ at the opposite electrode of a memristor, and the bottom panel illustrates the potential difference $u_{ij}$ across the memristor; a)  the waveform during the first phase resulting in the postive change of synaptic weight; b) the waveform during the second phase resulting in the negative change of synaptic weight (time units, frequency and pulse width are arbitrary).}
\end{figure}

Let us assume firstly that both of $x_i$ and $\delta_j$ are positive. Their values can be represented by two series of pulses, both with the amplitude $u_+/2$, so that pulses for $x_i$ an $\delta_j$ have opposite polarities and their number is proportional to the absolute values of the signals (Figure~ \ref{fig:pulsemodul}). With such representation, the voltage drop across the memristor exceeds the threshold $V_{on}$ only for those pulses, which simultaneously arrive to the opposite electrodes of the device.

According to this procedure, the weight change $\Delta w_{ij}$ is proportional to the minimum between 
$x_i$ and $\delta_j$
\begin{equation} 
	\Delta w_{ij} \propto\min(x_i, \delta_j).
	\label{eq:delta_as_min}
\end{equation}

In case $x_i$ and $\delta_j$ have opposite signs, $\Delta w_{ij}$ should decrease. This can be achieved by changing the pulse polarity. In the general case, when $x_i$ and $\delta_j$ can be either positive or negative, (\ref{eq:delta_as_min}) should be modified to:

\begin{equation}
	\Delta w_{ij} \propto sign(x_i\delta_j) \cdot \min(\left|x_i\right|, \left|\delta_j\right|)).
	\label{eq:delta_as_absmin}
\end{equation}

\begin{figure}[ht]
	\begin{center}
		\includegraphics[width=10cm]{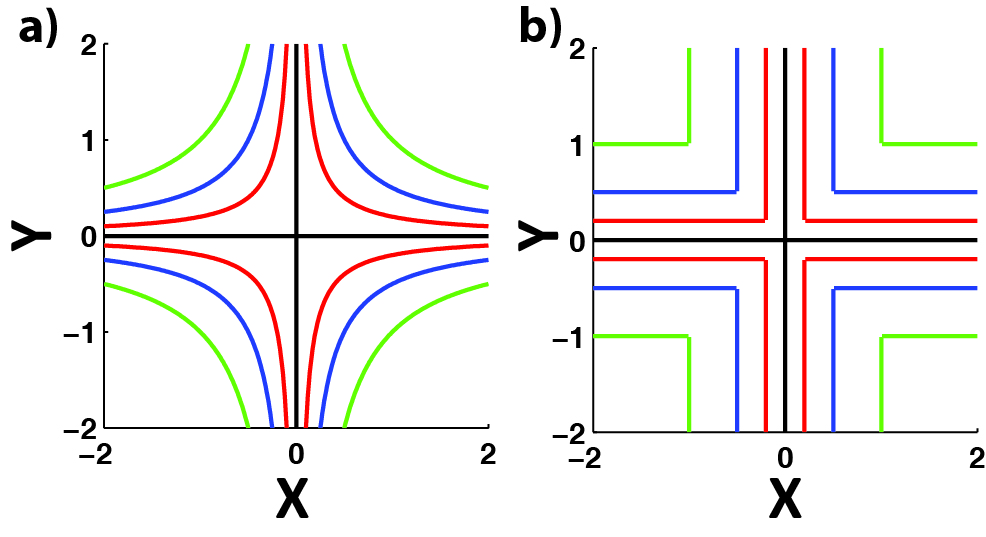}
	\end{center}
	\textbf{\refstepcounter{figure}\label{fig:multiplication_vs_min} Figure \arabic{figure}.}{ The similarity between landscapes of values for a product function $xy$ and $\sign(xy)\cdot\min(\left|x\right|, \left|y\right|)$ function; function value color notation increases from violet to blue and from yellow to red. }
\end{figure}

In fact, the expression (\ref{eq:delta_as_absmin}) approximates (\ref{eq:delta_as_product}) quite satisfactory for the network learning purposes (see Section 3 below), as it yields correct direction for a gradient descent process. The reason for this can be seen in Figure \ref{fig:multiplication_vs_min}. Therefore, this update rule will be used below in computational experiments.

In principle, the learning cycle should be divided into four subcycles for each combination of signs of $x_i$ and $\delta_j$. However, in this work we consider only non-negatively valued activation functions, and thus $x_i\geq$ 0. In this case, sign($x_i\cdot\delta_j$) depends only on $\delta_j$ and only two subcycles are required for the system to function.

The first subcycle for the positive value of sign($x_i\cdot\delta_j$) has been already described above. We further consider $u_-$ to be the voltage below the lower writing threshold $V_{off}$, i.e. by applying it on the memristor one decreases its conductivity, and $V_{off}<u_-$/2. In order to implement the second phase of the learning cycle (when $\Delta w_{ij}<0$), let us express the signals $x_i$ and $\delta_j$ in the form of pulses with an amplitude $|u_-|/2$, as shown in Figure~\ref{fig:pulsemodul}(b). As a result, the overall bias on the memristor is negative when pulses coincide, and the weight of memristor will decrease.

In summary, the learning procedure consists of two phases, which go successively one after another and are illustrated in Figure~\ref{fig:pulsemodul} (a) and (b). During the first phase, the $i$-th neuron sends a pulse train of positive polarity and amplitudes $|u_+|/2$ to the first electrode of the memristor, with the number of pulses proportional to the value of $x_i$. Simultaneously, in case of $\delta_j>0$ the $j$-th neuron sends a pulse train of negative polarity and amplitudes of $|u_+|/2$ to the second electrode of the memristor, with the number of pulses proportional to the value of $delta_j$, and the $j$-th neuron is inactive in case of $\delta_j<0$.

In the second phase, the $i$-th neuron sends a pulse train of negative polarity and amplitudes $|u_-|/2$ to the first electrode of the memristor, with the number of pulses proportional to the value of $x_i$. Simultaneously, in case of $\delta_j<0$ the $j$-th neuron sends to the second electrode of the memristor a pulse train of positive polarity and amplitudes of $|u_-|/2$, with the number of pulses proportional to the value of $\delta_j$, and the $j$-th neuron is inactive in case of $\delta_j>0$.

These two phases go sequentially one by one and form one full cycle of learning. It should be emphasized that both phases are always performed, irrespective of whether they have an effect or not. It is necessary that the neurons, located at the opposite ends of the memristor, work independently and rely only on the common schedule.

\section{Experiments}\label{Exper} 
All modelling experiments were performed in MATLAB.
In experiments, a standard two-layer network with [784-110-10] architecture was trained on the MNIST data set \citep{LeCun1998}. The MNIST dataset consists of 70,000 handwritten digits out of which 60,000 are used for the training process and 10,000 for the testing process. 

During the learning phase, we used minibatches of size 100 to speed up calculations. The initial weights were selected randomly from the uniform distribution, so that $w_{ij}^{(k)}\in[-a_k, a_k]$, where $a_k=1/\surd(n_{k-1})$, $n_{k-1}$ is the number of neurons in the $(k-1)$-th layer. 

At the beginning, we used the neuron sigmoidal activation function of the following type:
\begin{equation}
	x_i=f(z_i)=1/(1+e^{-z_i})
	\label{eq:sigm}
\end{equation}

In the further experiments, we used another neuron activation function called $relu$ (rectified linear unit):
\begin{equation}
	f(x)=\max\{0,x\}
	\label{eq:relu}
\end{equation}

The reason for the use of such activation function lies in the simplicity of its implementation in hardware. Despite the simplicity and the discontinuity of the derivative of (\ref{eq:relu}) at $x$=0, this function is widely used in machine learning and yields excellent results. For the weights update we used the equation (\ref{eq:delta_backprop}).

Three parameters, such as the method of multiplication, the matrix of error propagation, and the ``continuality'' or ``discreteness'' of variables in the implementation of equation (\ref{eq:delta_as_product}), were varied. Each of three described parameters has two possible options. In particular:

\begin{enumerate}
	\item As described in Section 2.3, two options of the multiplication method were examined. 
	First of all, we used a regular mathematical multiplication procedure in accordance with (\ref{eq:delta_as_product}). We refer to this procedure as the method $times$. The second option was to use the approximation (\ref{eq:delta_as_min}), the method called $absmin$.
	
	\item As described in Section 2.2, two options for the error propagation to the connection matrices were considered: (a) when the error is "back-propagated" across the same neuron connection weights, which works for the forward signal propagation (we called such standard method as $transposed$), and (b) when the error is conferred to the connection weights via the randomly chosen constant connection matrix, as has been specified in the equation (\ref{eq:delta_backprop_alignment}) and in \citep{Lillicrap2014}, which we called as $const$ method.
	
	\item Also, two cases in the implementation of (\ref{eq:delta_as_min}) were considered: in the first one the values of $x_i$ and $\delta_j$ are continuous, while in the second one these variables are made discrete before multiplication, as in the case illustrated in Figure~\ref{fig:pulsemodul}.
\end{enumerate} 

We assume that, when implemented in chip, the learning rate might depend on the frequency of pulses (see Figure~ \ref{fig:pulsemodul}) and therefore it can be easily varied by changing the pulse generator frequency in the whole network. Subsequently, the operation of changing the learning rate can be easily achieved. In the modelling experiments, starting from the value of $10^{-4}$, we dynamically change the learning rate depending on the training set error: if the error decreases, the epoch learning rate increases by $10\%$; if the error increases, the learning rate decreases by $30\%$. It is known that such dynamic change of the learning rate can speed up the  learning process. We have found that in the experiment the learning rate almost always stays in the range from $10^{-4}$ to $10^{-3}$. 

\begin{figure}[ht]
	\begin{center}
		\includegraphics[width=12cm]{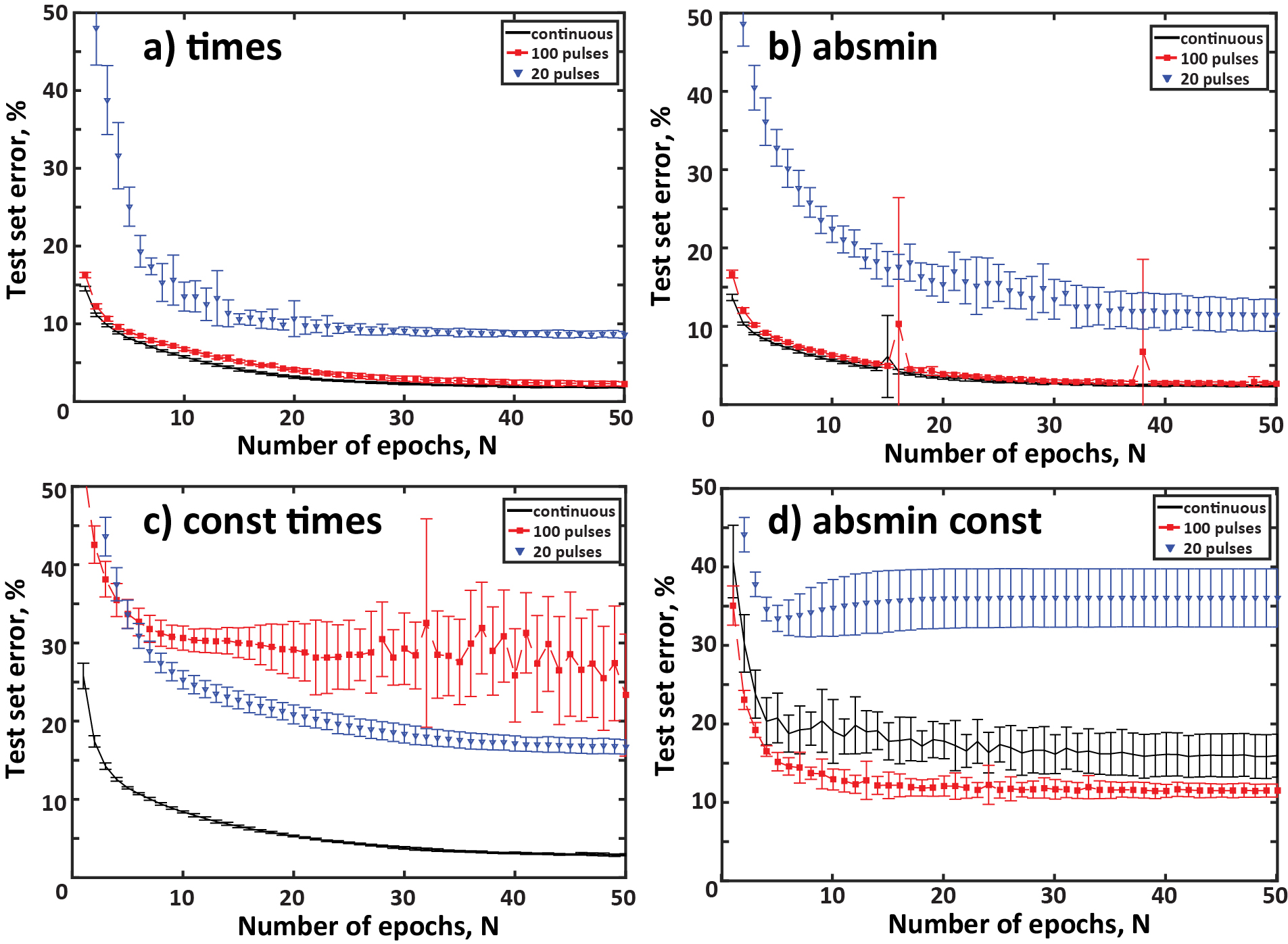}
	\end{center}
	\textbf{\refstepcounter{figure}\label{fig:relul} Figure \arabic{figure}.}{ Test set error for [[784-110-10]] network with relu-neurons and dynamically changing learning rate. The results for continuous values $x_i$ and $\delta_j$ aswell as for the discreteness of 100 and 20 pulses are shown with black solid lines, red dashed-lines and the blue triangles, respectively: a) method $times$, b) method $absmin$, c) method $times$ with $constant$ matrix, d) method $absmin$ with $constant$ matrix.}
\end{figure}

As was described above, the values of $x_i$ and $\delta_j$ are represented in the form of pulses, as shown in Figure~\ref{fig:pulsemodul}. However, in the modelling experiments, we have just discretized their values in the range from the experimentally found minimal to maximal values: $x_i\in [0, 5]$ and $\delta_j \in [-1, 2]$. The number of gradations can vary and in this paper we considered three cases: 20 and 100 gradations as well as a continuous one.

All the results were averaged by 10 trials.

\section{Results}\label{Results} 

The results are shown in Figure~\ref{fig:relul} and summarized in Table~\ref{Tab:01}. In the simulation  experiments, we used $relu$ (\ref{eq:relu}) activation function and dynamical adjustment of the learning rate.

\begin{table}[!t]
	\label{Tab:01} 
	\centering
	\caption{ Test set errors for various methods, after 50 epochs of learning}
	\begin{tabular}{lcccc}
		\toprule
		& \multicolumn{2}{c}{transposed} & \multicolumn{2}{c}{const}\\
		& times & absmin & times & absmin\\
		\midrule
		continuous & $1.8\% \pm 0.1\%$ & $2.4\% \pm 0.1\%$ & $2.9\% \pm 0.1\%$ & $16.0\% \pm 2.7 \%$\\
		100 pulses & $2.3\% \pm 0.3\%$ & $2.7\% \pm 0.1\%$ & $23.3\% \pm 7.8\%$ & $11.5\% \pm 0.8\%$ \\
		20 pulses & $8.6\% \pm 0.4\%$ & $11.4\% \pm 2.0\%$ & $16.7\% \pm 0.9\%$ & $36.0\% \pm 3.7\%$\\
		\bottomrule
	\end{tabular}
\end{table}

In the experiment, the test error smoothly decreases to about $1.8\%$ for standard backpropagation and to $2.4\%$ following the replacement of the $times$ with $absmin$ (black solid lines in Figure~\ref{fig:relul}). 

In fact, the results obtained with the method $times$ can be considered as standard and comparable with the state-of-the-art methods for this task. Significant improvements can be achieved only by increasing the number of hidden layer neurons and by increasing the training data set size with the use of elastic deformations of training examples (\citep{Meier2011}).

The results obtained for $x_i$ and $\delta_j$ discrete values are shown by dashed and dotted lines representing $\leq100$ and $\leq20$ pulses, respectively. One can see that given enough discrete levels (100 discrete values in our case) the performance is statistically close to that in a continuous case. However, the aggresive discretization leads to the rapid drop in the performance.

\section{Discussion}\label{Disc} 

The key issue addressed in our work is how to use the plasticity effects in synapses represented by memristors with multiple resistive states to locally implement the learning rule. The main distinction between our results and the related studies \citep{Nair2015} is that we have implemented the mechanism, which is able to propagate error backwards and is needed for multi-layered networks. This is important for the deep learning schemes. Note that in \citep{Nair2015} the single layer perceptron is considered, and the method proposed in that work cannot be used to propagate the error between the layers. The implementation of the learning rule requires the conversion of signals $x_i$ and $\delta_j$ at opposite electrodes of a memristor in a crossbar to the voltage drop across the crossbar that would change the memristor conductivity proportional to the product $x_i\times\delta_j$. We propose the mechanism based on the pulsed representation of signals $x_i$ and $\delta_j$ and implement the $absmin$ operation instead of the product. The use of $relu$ for the neuron transfer function also simplifies the implementation of neural networks as compared to the traditional sigmoidal transfer functions. Our results demonstrate that the memristor based implementation of error-based learning, including deep learning, is possible and can be efficient. 

The issues arising due to the thermal noise and the variability in memristors are still not resolved, however, the proposed scheme appears to be quite robust with respect to the introduction of switching errors below certain threshold, because the learning procedure does not require setting exact values on each iteration of the process --- the main requirement here is to move in the general direction of the error gradient. Hence, the reduction of the learning rate during such procedure should enable the system to eventually settle down in the desired minimum. The verification of this behavior can be performed using Monte-Carlo simulations, but they require at least sufficient statistical data on element performance or a good physical model, which is not available at the present moment. At the same time, the initial studies of the effect of noise as well as parameter variability on the operation of memristor-based schemes demonstrate their significant tolerance to these factors \citep{Nair2015}.  This fact speaks in support of our arguments presented above.

The important issue arising during the implementation of the backpropagation learning rule is the routing of the error signal $\delta$ in backward direction. In this paper, we have considered two possible approaches: the propagation via the same connections (same memristors), which are used in the forward direction, or, alternatively, the propagation via a bypass connection matrix which adopts random values and remains constant during the learning (method $const$). Despite the simplicity of the implementation on chip of the latter approach, our experiments have shown unsatisfactory results using such method. Nevertheless, due to the symmetry of the memristor crossbar, the error backpropagation can be done in the same way as a forward propagation during the inference step by using current summation.

The proposed replacement of the $times$ operation with $absmin$ turns out to give good results. The modified learning rule generally does point into the similar direction (preserving the sign) as the original gradient and yields relatively smooth performance curves (Figure~\ref{fig:relul}b). However, it also have some shortcomings. Firstly, the gradient calculation becomes inexact, which results in slightly larger  error ($2.4\%$ instead of $1.8\%$, see. Table~\ref{Tab:01}). Secondly, the large price to pay is the necessity of conversion of $x_i$ and $\delta_j$ into discrete pulses which seems laborious and decreases the accuracy for the small number of gradations. 

On the other hand, in such pulse representation, the memristor conductance changes by small steps proportionally to the number of pulses. Our approach is very close to that proposed by \citep{Soudry2014}, however, in our work we do not assume the linear dependence of the memristor conductance on the amplitude of the applied voltage. All pulses are assumed to be of the same amplitude, irrespective of the values of $x_i$ and $\delta_j$.

Finally, it should be emphasized that in this study we only demonstrate the operability of the proposed mechanisms at the proof of concept level. That is why we have restricted our work to the most commonly used benchmark--- the MNIST database, and have applied the number of hidden units, which is sufficient to show that the proposed simplifications still provide learning operation quality, surpassing that of "shallow" perceptron. The detailed studies of models of memristor-based deep learning systems and their physical implementation will follow. We believe that the approaches proposed in our work can expand the capabilities of memristor technologies in the  application for self-trained multilayer neural networks realization and decrease their power consumption by reducing the required number of memristors down to one-on-one connection weight. It is also worth noting that the techniques described in this paper are highly scalable.

\section*{Disclosure/Conflict-of-Interest Statement}
The authors declare that the research was conducted in the absence of any commercial or financial relationships that could be construed as a potential conflict of interest.

\section*{Acknowledgments}
\textit{Funding:} The work has been supported by Russian Science Foundation, Grant \# 14-19-01698.

\bibliographystyle{elsarticle-num}
\bibliography{Negrov_Manuscript}

\end{document}